\documentclass{bmvc2k}
\usepackage{multirow}
\usepackage{pifont}
\usepackage{multirow}
\usepackage{booktabs}
\usepackage{amsmath}
\usepackage{amssymb}
\usepackage{makecell}
\usepackage{adjustbox}

\title{Uni-Mlip: Unified Self-supervision for Medical Vision Language Pre-training}

\addauthor{Ameera Bawazir}{ameera.bawazir@tii.ae}{1}
\addauthor{Kebin Wu}{Kebin.wu@tii.ae}{1}
\addauthor{Wenbin Li}{wenbin.li@tii.ae}{1}

\addinstitution{
 Technology Innovation Institute\\
Abu Dhabi, UAE
}

\runninghead{Ameera Bawazir,Kebin Wu, and Wenbin Li}{Uni-Mlip}

\begin{document}

\maketitle

\begin{abstract}
Recent advancements in vision-language pre-training via contrastive learning have significantly improved performance across computer vision tasks. However, in the medical domain,  obtaining multimodal data is often costly and challenging due to privacy, sensitivity, and annotation complexity. To mitigate data scarcity while boosting model performance, we introduce \textbf{Uni-Mlip}, a unified self-supervision framework specifically designed to enhance medical vision-language pre-training. Uni-Mlip seamlessly integrates cross-modality, uni-modality, and fused-modality self-supervision techniques at the data-level and the feature-level. Additionally, Uni-Mlip tailors uni-modal image self-supervision to accommodate the unique characteristics of medical images. Our experiments across datasets of varying scales demonstrate that Uni-Mlip significantly surpasses current state-of-the-art methods in three key downstream tasks: image-text retrieval, image classification, and visual question answering (VQA).
\end{abstract}

\section{Introduction (Uni-Mlip)}
\label{sec:intro}
Vision-and-Language Pre-training (VLP) techniques, such as the Contrastive Language Image Pre-training (CLIP) model~\cite{Radford2021LearningTV}, have set a foundational approach for leveraging self-supervision with language guidance to integrate visual and textual data effectively. By aligning image and text representations through contrastive learning, CLIP improves the model's ability to interpret complex data by linking visual concepts to linguistic descriptions. This method significantly enhances the understanding of multimodal representations by pre-training on extensive datasets and subsequently fine-tuning on targeted downstream tasks. 
This paradigm shift is particularly vital in the medical domain, where the acquisition of multimodal medical data presents significant challenges due to concerns over data privacy, sensitivity, and the complex, domain-specific knowledge required for annotation.

Medical Vision-and-Language Pre-training (Med-VLP) aims to address such challenges inherent to the medical imaging field. The exploration of self-supervised learning (SSL) techniques in medical VLP, as seen in models like ConVIRT \cite{zhang2022contrastive} and GLoRIA \cite{Huang_2021_ICCV}, highlight the shift towards leveraging unlabelled data, indicating a growing emphasis on models that can learn from the nuanced and complex nature of medical images and texts without extensive manual annotations. Despite advancements, previous approaches still lack the ability to systematically integrate and contextualize medical-specific knowledge, which leads to limited multimodal understanding and contextual reasoning. 

In response to this, our paper introduces \textbf{Uni-Mlip}, a novel \textbf{Uni}fied \textbf{M}edical C\textbf{LIP} framework designed to improve medical vision-language pre-training with enhanced self-supervision. Uni-Mlip seamlessly integrates cross-modal, uni-modal, and fused-modal self-supervision techniques at both data-level and feature-level. This approach allows for more effective and nuanced mining of medical data, facilitating the learning of aligned and transferable features that excel in downstream tasks. Our main contributions include: \textbf{(1)} Uni-Mlip is the first unified framework that systematically explores feature-level and data-level self-supervision across both uni-modality and multimodality settings to cohesively align image and text modalities, to the best of our knowledge; \textbf{(2)} Uni-Mlip adapts image self-supervision techniques and tailors them to the unique requirements for higher precision and detail sensitivity characteristic~\cite{article} of medical images; \textbf{(3)} Comprehensive experiments demonstrate Uni-Mlip's superiority over current state-of-the-art methods, showcasing its effectiveness in tasks including image-text retrieval, image classification, and visual question answering (VQA). 

\section{Related work}
The CLIP model \cite{Radford2021LearningTV} pioneers vision-language pre-training by aligning image-text pairs by maximizing the similarity between image and text features in a shared latent space. Subsequent models build upon CLIP by enhancing learning objectives \cite{9878693uni, Yao2021FILIPFI}, modifying pre-training architecture \cite{10204385clippo, articleunifiying}, and expanding downstream tasks \cite{9879567GLIP, NEURIPS2022_3ba96055}, which allow for better representation learning and widen the scope of vision-language pre-training.
\\
\textbf{Medical Vision-and-Language Pre-Training (Med-VLP)}. 
The models above optimized for natural images often fall short in medical settings: medical images necessitate a meticulous focus on fine-grained details, and critical diagnostic information is often embedded in the absolute pixel intensity, the scale of abnormalities, and their precise locations within the image. Hence, domain-specific Med-VLPs have been proposed to overcome these challenges. Medical VLP models such as GLoRIA \cite{9710099gloria}, LoVT \cite{muller2022joint}, and PRIOR \cite{cheng2023prior} introduced global-local image-text alignment, enabling the learned models to capture fine-grained information lying in the medical images. PMC-CLIP \cite{lin2023pmc} presents a CLIP-like model architecture that is trained on a large-scale dataset of medical image-caption pairs, highlighting the critical role of extensive datasets in effective pre-training. A significant contribution of their work is the introduction of PMC-OA, a newly curated dataset of image-caption pairs sourced from biomedical literature available on PubMedCentral’s OpenAccess. \\
\textbf{Self-Supervised learning (SSL)}. Self-Supervised Learning~\cite{caron2021emerging, grill2020bootstrap, chen2020simple} addresses the issue of data annotation scarcity by training models to learn meaningful representations with unlabelled data. For natural images, adding traditional SSL to the CLIP framework for joint training can significantly improve the performance, whereas the study in \cite{huang2024radiology, Mu2022Slip} found the opposite when directly trained on medical images. Firstly, the inherent assumptions of conventional SSL, i.e., image augmentations do not alter the semantic information of an image, may not hold when applied to medical images. This requires careful adaptation of augmentations to maintain the diagnostic integrity of the image.
Secondly, conventional SSL generally applies augmentation techniques to the input space only. We argue the feature-level SSL, through perturbation in the representation space, can further boost the performance and lead to more robust and more discriminative learning results. 
To this end, other approaches such as M3AE \cite{10.1007/978-3-031-16443-9_65} and MUMC \cite{10.1007/978-3-031-43907-0_36} respectively added a self-supervised learning paradigm to enhance the Med-VLP performance. M3AE reconstructed missing pixels and tokens in randomly masked images and texts using visual and textual features from different layers; MUMC utilized momentum contrastive self-supervised approach for uni-modal and multimodal representations learning. However, despite achieving state-of-the-art performance on the VQA task, both M3AE and MUMC have notable limitations.
Firstly, MUMC is specifically designed for the VQA task, limiting its generalizability to other tasks. Secondly, M3AE and MUMC integrate several complex components, including heavy multimodal fusion modules, to create intensive interactions between image and text modalities. While this complexity enhances performance on VQA, it does not translate as effectively to other tasks, such as image-text retrieval, where M3AE, in particular, underperforms. Additionally, both models rely on large model sizes for training and inference. The integration of additional components on top of the pre-trained encoders during the inference stage, instead of adding a simple classification layer, significantly increases the model parameters. Moreover, the reliance on masked inputs for both modalities in M3AE and MUMC can potentially weaken their zero-shot capabilities when evaluated on unmasked data.

In contrast, our proposed method, Uni-Mlip, aims to provide a general-purpose Med-VLP model capable of consistently achieving strong performance across various downstream tasks. It does so by incorporating both data-level and feature-level SSL in cross-modality, uni-modality, and fused-modality settings, tailored specifically to medical data, while maintaining the simplicity of the original CLIP architecture.
\section{Uni-Mlip model}
In this section, we present a comprehensive overview of our proposed methodology Uni-Mlip, encompassing a detail explanation of our model architecture and pre-training objectives. 
\begin{figure*}
\begin{center}
\includegraphics[width=0.95\textwidth]{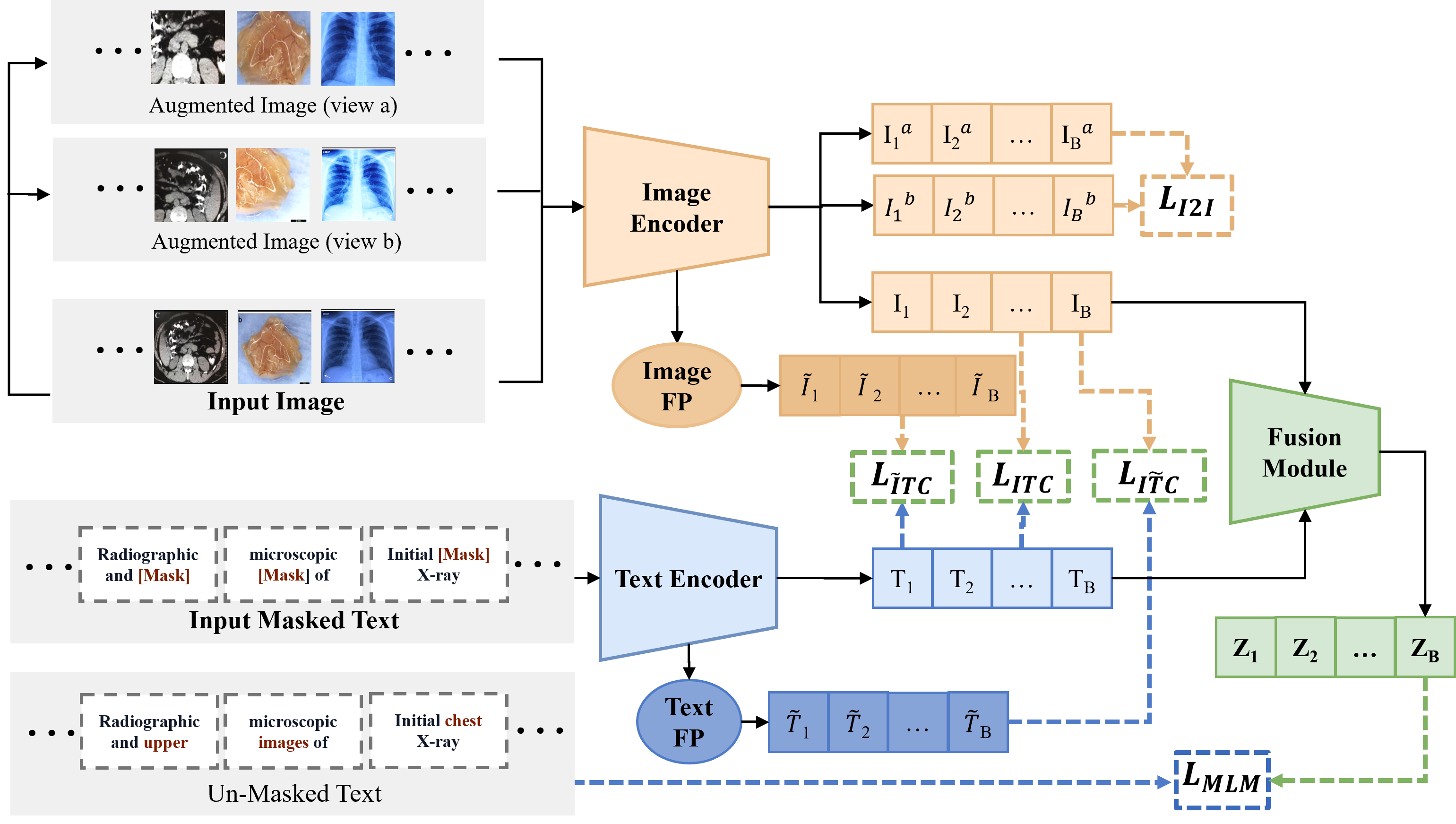}
\end{center}
   \caption{Illustration of our proposed model architecture Uni-Mlip. The colors indicate modality-specific components such that orange is for image, blue is for text, and green is for multimodal based components.}
\label{fig:architecture}
\end{figure*}
As shown in Figure \ref{fig:architecture}, in our pre-training model design, we follow a CLIP-like architecture that separately encodes each modality with a modality-specific encoder, and then jointly optimize both of them through image-text contrastive ($\textbf{ITC}$) loss, ensuring an alignment of image and text embedding. Let V be an input image, and C be the corresponding caption,  given a batch of image-caption pairs {$(V_1,C_1)$,....,$(V_B, C_B)$}, where $V_i$ $\in \mathbb{R}^{H \times W \times C}$ is the $i$-{th} image and $C_i$ is the paired caption, we encode the input image via the image encoder $E_{V}$ and the input text via the text encoder $E_{T}$ to get the image and text features respectively. These features are then projected to a shared embedding space and normalized to get the final image embedding $I\in \mathbb{R}^{B \times D}$ and text embeddings $T\in \mathbb{R}^{B \times D}$, where $B$ is the batch size and $D$ is the shared embedding dimension.   

In Uni-Mlip, we integrate several complementary Self-Supervised Learning (SSL) techniques into the architecture reminiscent of CLIP. This involves incorporating cross-modality SSL at both the input and feature levels to align the embedding of images and text. Additionally, we introduce fused-modal and uni-modal SSL for text and images respectively to enhance robustness even further. 

\subsection{Cross-modal input level self-supervision }
We perform an input level SSL by contrasting the image and text embeddings using the Image-Text Contrastive (ITC) loss from CLIP~\cite{Radford2021LearningTV}. The ITC loss is defined as a sum of image-to-text $L_{I2T}$ and text-to-image  $L_{T2I}$ NT-Xent losses \cite{NIPS2016_6b180037}, which is formulated as:
\begin{equation}
\begin{split}
\mathcal{L}_{ITC}  &= -\frac{1}{2B} \sum_{i=1}^{B} log \frac{e^{sim(I_{i}, T_{i})/ \tau} }{\sum_{j=1}^{B} e^{sim(I_{i}, T_{j})/ \tau} }  -\frac{1}{2B} \sum_{i=1}^{B} log \frac{e^{sim(I_{i}, T_{i})/ \tau} }{\sum_{j=1}^{B} e^{sim(I_{j}, T_{i})/ \tau} }
\end{split}
\end{equation}
The goal of the ITC loss is to maximize the similarity between the embedding of the B image-text pairs i.e.,$(I_{i}, T_{i})$ while minimizing the similarity with the rest of the $B^2 - B$  non-pair samples i.e.,$(I_{j}, T_{i})$ in the batch, where $i\neq j$. The temperature  $\tau$ is a learnable parameter for scaling the measured similarity. The image embedding $I$ and text embedding $T$ are obtained via the image and text encoder, and $sim$ is the cosine similarity metric.  
\subsection{Cross-modal feature level self-supervision}
To further boost the cross-modality alignment, we also introduce a feature-level self-supervision by performing a feature perturbation. The motivation is that image/text features after some minor perturbations are expected to remain semantically aligned with the corresponding pair of text/image embeddings, ensuring robust and better representation learning. The feature perturbation can be implemented by a simple Dropout \cite{Hinton2012ImprovingNN} or its variant such as DropBlock \cite{ghiasi2018dropblock}. Perturbation applied to the image features is done after the last residual layer and before the attention pooling layer in the vision encoder, where a DropBlock method is utilized. On the other hand, to generate the perturbed text features, a dropout is applied after getting the text features from the text encoder. In Figure \ref{fig:architecture}, we refer to the DropBlock operation as Image FP, and to the dropout operation as Text FP. 

Regarding the training objectives, we align the image/text features with the perturbed features of text/image using contrastive learning. Hence, the ITC loss with perturbed image features is denoted as $\mathcal{L}_{\tilde{I}TC}$,  where the image embedding $I$ from $\mathcal{L}_{ITC}$ is replaced with the perturbed image embedding $\tilde{I}$. Similarly, we define the ITC loss with the perturbed text embeddings as $\mathcal{L}_{I\tilde{T}C}$, such that the text embedding $T$ in $\mathcal{L}_{ITC}$ equation is replaced with perturbed text embedding $\tilde{T}$.

\subsection{Uni-modal self supervision - image}\label{sec:simclr}

Based on the observation~\cite{Mu2022Slip} that combining image-only self-supervised learning with language supervision can effectively improve the representation quality of the images, we move one step further and add uni-modal, image-only or text-only, self-supervision on top of cross-modal supervision. While reconstructing masked images is a popular approach for natural images {\cite{he2022masked}, this method may be less suitable for medical imaging. This is because medical images often depict localized anomalies associated with diseases, and masking these parts could hinder the model's ability to accurately represent these localized anomalies. Thus, we adopt the contrastive self-supervised learning objective from SimCLR~\cite{chen2020simple}, which enables better capturing of subtle patterns and anomalies in medical images.

Given a batch size $B$ of images, we create two strongly augmented views for each image. Then the uni-modal contrastive loss $\mathcal{L}_{I2I}$ is defined in Eq.~\ref{eq:li2i} to maximize the similarity between the embedding of the positive pairs (views of the same image) while minimizing the similarity between the negative pairs (views of different images). Such loss term promotes the model to learn semantically robust representations for images with different augmentations.
\begin{equation}\label{eq:li2i}
\begin{split}
\mathcal{L}_{I2I} & = -\frac{1}{2B} \sum_{i=1}^{B} log \frac{e^{sim(I_{i}^{a}, I_{i}^{b})/ \tau} }{\sum_{j=1}^{B} e^{sim(I_{i}^{a}, I_{j}^{b})/ \tau} } -\frac{1}{2B} \sum_{i=1}^{B} log \frac{e^{sim(I_{i}^{a}, I_{i}^{b})/ \tau} }{\sum_{j=1}^{B} e^{sim(I_{j}^{a}, I_{i}^{b})/ \tau} }
\end{split}
\end{equation}
where $I_{i}^{a}$ is the feature embedding of the augmented image $V^{a}$, and $I_{i}^{b}$ is the feature embedding of the augmented image $V^{b}$. 

While SLIP \cite{Mu2022Slip} empirically demonstrates SimCLR's enhancement of vision-language pre-training with natural images, simply adding Eq. \ref{eq:li2i} worsens the CLIP performance in the medical domain \cite{huang2024radiology}. One reason may be the fundamental differences in characteristics between natural and medical images. Unlike natural images, for which relative intensity information is key in recognition, medical images rely heavily on precise absolute intensity values. When integrating SimCLR into CLIP-like architecture, the same vision encoder will process three different views for each input image: one weakly augmented view used for CLIP, and two views with strong augmentation for SimCLR. When the vision encoder, such as a ResNet, includes batch normalization layers, strong image augmentations will significantly alter the mean and variance statistics in batch normalization layers, harming the intensity information critical for medical images. To address this issue, we propose to train without SimCLR for some epochs first, and then we freeze the mean and variance in the batch normalization layers while keeping the weight and bias as learnable parameters when incorporating the SimCLR loss (Eq. \ref{eq:li2i}). This approach ensures that strong augmentations introduced by SimCLR will not alter the mean and variance in batch normalization layers, allowing for better vision representation learning for medical images.  Experiments in Section \ref{sec:ablation} support our analysis and prove the effectiveness of our proposed solution.

\subsection{ Fused-modal self-supervision}

To enhance text representation, we integrate fused-modal self-supervision for the textual modality. Motivated by MMBERT \cite{khare2021mmbert} and PMC-CLIP \cite{lin2023pmc}, we utilize masked language modeling (MLM) self-supervision task for the input captions. In masked language modeling, the model predicts the masked text token by learning from the medical text features while using the image feature as a supplementary context. Before being fed to the text encoder, the captions are masked randomly with a probability of $15\%$. Following BERT \cite{Devlin2019BERTPO}, our masked tokens are replaced with a unique token [MASK]. After that, the visual and masked linguistic features obtained from the image and text encoder are fused by a multimodal transformer-based fusion module. The fusion module generates the full caption by replacing the [mask] token with the predicted text token. Hence the MLM loss is defined as:

\begin{equation}
\mathcal{L}_{MLM} = {\mathbb{E}}_{(V,C) \sim D}[CE(y^{mask}, p^{mask}(V,C)]
\end{equation}

where $y^{mask}$ denotes the ground truth and $p^{mask}$ is the predicted masked token generated by the fusion module. The masked language modeling objective enables the model to focus on learning fine-grained details from the medical text to improve the representation and understanding of language-specific information. 

\subsection{Total training loss}
The total loss is a weighted sum of all of the individual cross-modal, uni-modal and fused-modal losses and therefore is formulated as: 
\begin{equation}
\begin{split}
\mathcal{L}_{Total}  = \lambda_{cm} \cdot (\mathcal{L}_{ITC} + \mathcal{L}_{\tilde{I}TC} + \mathcal{L}_{I\tilde{T}C}) + \lambda_{um} \cdot ( \mathcal{L}_{I2I}) + \lambda_{fm} \cdot (\mathcal{L}_{MLM})
\end{split}
\end{equation}
where $\lambda_{cm}$ , $\lambda_{um}$, and $\lambda_{fm}$ refer to the weight value assigned to the cross-modal, uni-modal, and fused-modal losses respectively during training.  

\section{Experiment settings}
 We  pre-train the Uni-Mlip on a large-scale dataset and then evaluate it on three downstream tasks. We provide the dataset and implementation details below. 

\subsection{Dataset}
For pre-training, we use PMC-OA \cite{lin2023pmc}, a vast open-source medical image-caption dataset comprising 1.65M pairs sourced from PubMedCentral's OpenAccess. 

For image-text retrieval evaluation, including both image-to-text (I2T) and text-to-image (T2I) retrieval tasks, we employ the Radiology Objects in COntext (ROCO) dataset \cite{10.1007/978-3-030-01364-6_20}, where 2000 samples are randomly selected in evaluation following \cite{lin2023pmc, subramanian2020medicat, 10.1007/978-3-031-16443-9_65}. As there is no overlapping between ROCO and PMC-OA, such a task also serves as a zero-shot evaluation for Uni-Mlip. For classification, we assess our model on three datasets: NIH \cite{Wang_2017}, CheXpert \cite{articlechexpert} and MIMIC \cite{Johnson2019}, consisting of 112,120, 224,316, and 377,110 chest X-ray images respectively with various disease labels. Additionally, for the downstream task VQA, we use VQA-RAD \cite{Lau2018} with 315 images and 3,515 questions, and SLAKE \cite{Liu2021} with 642 images and 14,028 questions, both including closed-ended and open-ended questions on radiology images.
\subsection{Implementation details}
\textbf{Implementation Details of Model Pre-training}. 
Following \cite{lin2023pmc}, we adopt a modified ResNet50 \cite{he2016deep} architecture as a vision encoder $E_V$, and PubMedBERT \cite{gu2021domain} as our text encoder $E_T$. Our fusion module employs four transformer layers to concatenate image and text embedding. Perturbation applied to image features uses DropBlock \cite{ghiasi2018dropblock} with a 50$\%$ probability with a block size of 3x3, while perturbation on text features are carried out by using dropout \cite{Hinton2012ImprovingNN} with a 75$\%$ probability. For the image-only self-supervision branch, the implementation of the SimCLR \cite{chen2020simple} augmentation includes color jittering, random gray-scale conversion, random Gaussian blurring, and horizontal flipping. 
All images are resized to 224x224 and trained with a batch size of 768 across 8 NVIDIA V100 GPUs for 100 epochs. The training employs a learning rate of 1e-4, with cosine scheduling and AdamW optimizer. For the training loss weights,  we set $\lambda_{cm} = \frac{0.5}{3} = 0.167$, $\lambda_{um} = 0.5$, and $\lambda_{fm} = 0.5$ as default values, which are empirically determined.

As discussed in Section \ref{sec:simclr}, freezing batch normalization is required to incorporate image-only self-supervision into the framework. Therefore, we initialize the three primary components—vision encoder, text encoder, and fusion module—with pre-trained weights from the 100th epoch of the PMC-CLIP model and freeze the mean and variance in the batch normalization layers during subsequent training. To ensure a fair comparison with PMC-CLIP, we re-implement PMC-CLIP by extending the model training to 200 epochs, thus doubling the original duration while retaining identical hyper-parameters. \\
\textbf{Implementation Details of Downstream Tasks}. For cross-modal retrieval, we randomly select 2000 test image-text pairs following \cite{lin2023pmc, 10.1007/978-3-031-16443-9_65} and resize the images to 224x224. For image-to-text retrieval, we calculate the inner product between the image feature and each text feature in the test set. The text with the highest inner product value is deemed the predicted text label for the image. The recall is then computed for the top 1, 5, and 10 captions. The text-to-image retrieval process is similar.

Regarding the classification task, we follow \cite{seyyed2020chexclusion} for dataset splits to mitigate comparison bias. With the pre-trained modified ResNet50 in Uni-Mlip, we add a linear classification layer on top for predictions. Input images are resized to 224x224, with batch sizes of 190 for CheXpert, and 150 for MIMIC and NIH. We perform end-to-end fine-tuning with binary cross-entropy loss and Adam optimizer, and the training is up to 64 epochs with early stopping criteria. The performance is evaluated with the macro Area Under the ROC Curve (AUC) metric.

For visual question answering (VQA), we cast the VQA as a classification task following the works in \cite{10.1007/978-3-031-16443-9_65} and \cite{dou2022empirical}. Instead of appending lots of new self-attention layers as in \cite{10.1007/978-3-031-16443-9_65} and \cite{dou2022empirical} to fuse the vision and text representations, we simplify the model framework with four components: a vision encoder, a text encoder, a one-layer transformer to fuse vision and text features, and a two-layer MLP to serve as the classification head. In addition, we append 20 learnable tokens in encoding questions, which can improve the representation power of text embedding effectively. The vision and text encoders are initialized with our pre-trained models and then trained with a learning rate of 5e-5. Meanwhile, the fusion and classification modules are randomly initialized and trained with a higher rate of 5e-4. 

\section{Empirical results}
In this section, we quantitatively demonstrate Uni-Mlip's superiority in image-text retrieval, image classification, and VQA, and provide comprehensive ablation studies to highlight the effectiveness of each component in our unified self-supervised framework.

\subsection{Downstream task results}
\begin{figure*}[tp]
  \centering
  \includegraphics[width=1\textwidth]{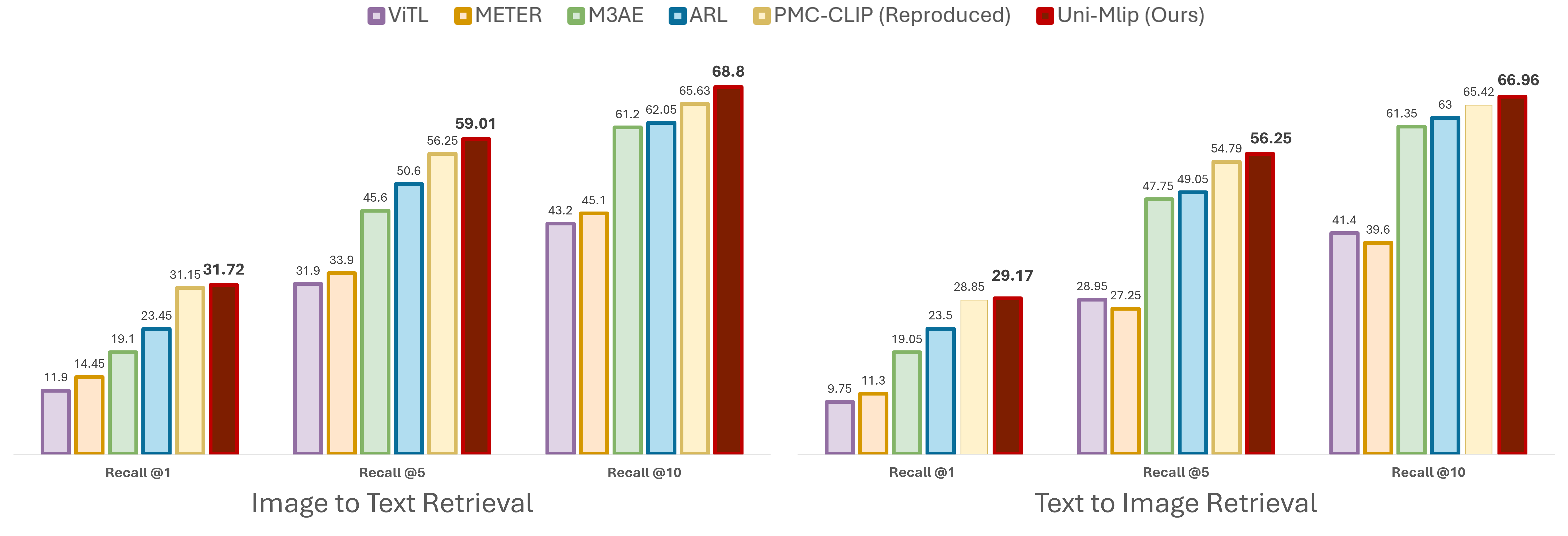} 
  \caption{Evaluation Results on Image-to-Text (left) and Text-to-Image (right) retrieval tasks on ROCO dataset. }
  \label{fig:retreival results}
\end{figure*}
\textbf{Image-Text Retrieval} For image-text retrieval, we compare our results with ViTL\cite{kim2021vilt}, and METER \cite{dou2022empirical}, M3AE \cite{10.1007/978-3-031-16443-9_65}, ARL \cite{chen2022align}, and our own reproduction of PMC-CLIP \cite{lin2023pmc}. As shown in Figure \ref{fig:retreival results}, Uni-Mlip consistently outperforms the others under both image-to-text and text-to-image tasks across Recall@1, Recall@5, and Recall@10 metrics. In particular, in image-to-text retrieval, Uni-Mlip surpasses the previous state-of-the-art by 3.17$\%$  in Recall@10 and achieves a 2.76$\%$ improvement in Recall@5. Similarly, in text-to-image retrieval, the model outperforms PMC-CLIP with gains of 1.46$\%$ in Recall@5 and 1.54$\%$ in Recall@10.This task, regarded as a zero-shot evaluation, shows that our method with cross-modality, uni-modality, and fused-modality SSL objectives at both the input and feature levels achieves superior alignment between medical image and text, facilitating better retrieval performance. \\
\textbf{Medical Image Classification}
For image classification, we compare the performance of our solution with previous state-of-the-art medical vision-language models. As shown from Table \ref{tab:classification}, the  AUC scores indicate that Uni-Mlip achieves the best results for different data splits  (1$\%$, 10$\%$, and 100$\%$) on MIMIC \cite{Johnson2019}, CXP \cite{articlechexpert} and NIH \cite{Wang_2017} datasets. Notably, when using 10$\%$ of the data, Uni-Mlip achieves an AUC of 79.1$\%$ on MIMIC, a gain of 1.7$\%$ over the previous best performance of 77.41$\%$ by PMC-CLIP. Our model consistently outperforms baselines that do not employ image uni-modal self-supervision, which further validates the importance of incorporating uni-modal image self-supervision to improve the visual representations, resulting in a better performance on medical image classification tasks.  \\
\textbf{Medical Visual Question Answering} For the VQA task, we report each dataset's open-ended, close-ended, and overall accuracy. Table \ref{tab:vqa} demonstrates the superiority of Uni-Mlip in the overall evaluations. Moreover, Uni-Mlip gains consistently on closed-ended VQA with substantial improvement of 1.48$\%$ on VQA-RAD and 1.44 $\%$ on Slake over PMC-CLIP. Hence, Uni-Mlip excels at leveraging medical text and image data to answer visual questions accurately.

While cross-modal and fused-modal interactions are crucial for integrating visual and textual information, uni-modal supervision provides a foundation for robust modality-specific representations. Our Uni-Mlip effectively leverages the benefits of cross-modal, uni-modal, and fused-modal self-supervision, enhancing the overall representation quality and task-specific performance. In addition, the superior performances of Uni-Mlip in the tasks of image classification and VQA, where finetuning is performed after initializing with the pre-trained Uni-Mlip,  further indicate that the proposed framework also offers a robust starting point in the optimization space suitable for other task-oriented fine-tuning. 

\begin{table}[tp]
\begin{center}
\begin{adjustbox}{max width=\textwidth}
\begin{tabular}{|l|ccc|ccc|ccc|}
\hline
\multirow{2}{*}{\textbf{Method}} & \multicolumn{3}{c|}{\textbf{MIMIC}} & \multicolumn{3}{c|}{\textbf{CXP}} & \multicolumn{3}{c|}{\textbf{NIH}} \\ \cline{2-10} 
 & 1\%     & 10\%    & 100\%  & 1\%     & 10\%    & 100\%   & 1\%     & 10\%    & 100\%   \\ \hline
Random   &53.6 & 66.5 & 78.2  & 62.6    & 69.0    & 76.9    & 56.4    & 67.1    & 76.9    \\
ImageNet & 67.8 &70.5 &79.3 & 63.7    & 70.7    & 77.7    & 59.7    & 68.9    & 78.1    \\ \hline
ConVIRT \cite{zhang2022contrastive} &67.8 &73.4 &80.1  & 63.2    & 71.3    & 77.7    & 60.0    & 69.0    & 76.6    \\
GLoRIA \cite{Huang_2021_ICCV} &67.5 &72.6 &80.1    & 62.9    & 69.0    & 77.8    & 60.1    & 71.2    & 77.7    \\
MGCA \cite{wang2022multigranularity} &68.4 &74.4 &80.2    & 63.4    & 72.1    & 78.1    & 61.1    & 67.8    & 77.3    \\
M-FLAG \cite{liu2023mflag} & 69.5 &74.8 &80.2  & 64.4    & 71.4    & 78.1    & 62.2    & 71.6    & 78.7    \\ 
PMC-CLIP \cite{lin2023pmc} (reproduced) & 73.1 &77.4 &81.8  & 69.1   & 74.9   & 79.1    & 64.9   & 76.3  & 82.3    \\ 
Uni-Mlip (ours) &\textbf{73.2} &\textbf{79.1} & \textbf{82.0}  &\textbf{69.1}   & \textbf{75.3}   & \textbf{79.8}    & \textbf{65.6}   & \textbf{76.4}  & \textbf{82.9}    \\ \hline
\end{tabular}
\end{adjustbox}
\end{center}
\caption{Classification results on MIMIC, CXP and NIH dataset. }
\label{tab:classification}
\end{table}

\begin{table*}[tp]
\begin{center}
\begin{tabular}{|l|ccc|ccc|}
\hline
\multirow{2}{*}{\textbf{Methods}} & \multicolumn{3}{c|}{\textbf{VQA-RAD}} & \multicolumn{3}{c|}{\textbf{Slake}} \\ 
\cline{2-7} 
  
                         & Open    & Closed  & Overall   & Open    & Closed  & Overall    \\ \hline
MEVF-BAN \cite{nguyen2019overcoming}                   & 49.20    & 77.20    & 66.10    & 77.80    & 79.80    & 78.60  \\
CPRD-BAN \cite{10.1007/978-3-030-87196-3_20}               & 52.50    & 77.90    & 67.80    & 79.50    & 83.40    & 81.10   \\
PubMedCLIP \cite{eslami2021does}             & 60.10    & 80.00    & 72.10    & 78.40    & 82.50    & 80.10   \\
PMC-CLIP \cite{lin2023pmc} (reproduced)                 & \textbf{61.45}    & 80.14    & 72.73    & \textbf{80.16}   &  84.38   &  81.81 \\ 
Uni-Mlip (ours)         & 60.43    & \textbf{81.62}    & \textbf{73.17}   & 79.38    &  \textbf{85.82}  &  \textbf{81.90}  \\ \hline
\end{tabular}
\end{center}
\caption{VQA results on VQA-RAD and Slake}
\label{tab:vqa}
\end{table*}

\subsection{Ablation studies}\label{sec:ablation}
To accelerate our analysis and reduce computational demands, we conducted ablation experiments using the ROCO dataset \cite{10.1007/978-3-030-01364-6_20} instead of PMC-OA for pre-training, focusing on image-text retrieval tasks. Performance evaluation was conducted on both image-to-text (I2T) and text-to-image (T2I) retrieval tasks, reporting the top 1 recall value.

In Table \ref{tab:Ablation_BN}, we illustrate the impact of freezing the mean and variance in the batch normalization layer when integrating SimCLR into the CLIP-based model for medical data. Consistently, V represents the input image, while $V^{a}$ and $V^{b}$ depict strongly augmented images used in SimCLR. Notably, the second row reveals a significant performance drop upon the naive addition of SimCLR, consistent with the observation in \cite{huang2024radiology}. We further analyze this by feeding the augmented images in the forward pass while setting the weight of $L_{I2I}$ loss to zero (third row). The still low performance reveals that the drop is due to feeding strongly augmented images in the forward pass. Consequently, we propose freezing batch statistics in batch normalization layers when incorporating the image-only SSL loss $L_{I2I}$, resulting in a noticeable improvement over the baseline as shown in the last row. This study emphasizes the importance of freezing batch normalization when integrating the SimCLR framework for Med-VLP.

Moreover, we demonstrate the effectiveness of our method by evaluating the significance of each learning objective. As demonstrated in Table \ref{Ablation lrobjectives}, incorporating cross-modal self-supervision at the feature level yields notable performance gains of 1.0 and 1.6 in the I2T retrieval task, and 1.6 and 1.7 in the T2I retrieval task for $L_{I\tilde{T}C}$ and $L_{\tilde{I}TC}$, respectively. Additionally, integrating the $L_{I2I}$ objective significantly boosts performance by 3.5 and 2.3 in the I2T and T2I tasks, respectively. Ultimately, the combination of all five training objectives results in superior performance across both tasks, highlighting the significance of our unified framework with cross-modal self-supervisions, both input-level and feature-level, as well as uni-modal and fused-modal self-supervisions.

\begin{table}[tp]
\begin{center}
\begin{tabular}{cccccccccc}
\toprule
Vision Encoder Input  & Training Objectives  & Freeze BN & I2T  & T2I  \\
\midrule
$V$ & $L_{ITC} , L_{MLM}$ & \ding{55}  & 22.2 & 21.7  \\
$V, V^{a}, V^{b}$ & $L_{ITC} , L_{MLM} , L_{I2I}$ & \ding{55}  & 5.0 & 1.0  \\
$V, V^{a}, V^{b}$ & $L_{ITC} , L_{MLM} $ & \ding{55}  & 4.0 & 6.9  \\
$V, V^{a}, V^{b}$ & $L_{ITC} , L_{MLM} , L_{I2I}$ & \ding{51}  & 25.7 & 24.0  \\

\bottomrule
\end{tabular}
\end{center}
\caption{Ablation study on ROCO dataset showing the importance of freezing batch normalization when incorporating image-to-image self-supervised learning loss $L_{I2I}$.}
\label{tab:Ablation_BN}
\end{table}

\begin{table}[tp]
\begin{center}
\begin{tabular}{ccccccccccccc}
\toprule
$L_{ITC}$ & $L_{MLM}$  & $L_{I\tilde{T}C}$ & $L_{\tilde{I}TC}$ & $L_{I2I}$ & I2T  & T2I & Performance Gain (I2T/T2I) \\
\midrule
 \ding{51} & \ding{51} & \ding{55} & \ding{55} & \ding{55} & 22.2 & 21.7 &  0.0 / 0.0 \\
 \ding{51} & \ding{51} & \ding{51}  & \ding{55} & \ding{55} & 23.2 & 23.3 & +1.0  / +1.6   \\
\ding{51} & \ding{51} & \ding{55}  & \ding{51} & \ding{55} & 23.8  & 23.4 &  +1.6 / +1.7 \\
\ding{51} & \ding{51} & \ding{55}  & \ding{55} & \ding{51} & 25.7 & 24.0  & +3.5 / +2.3 \\
\ding{51} & \ding{51} & \ding{51} & \ding{51} & \ding{51} & \textbf{25.9} & \textbf{24.5} & +3.7 / + 2.8\\
\bottomrule
\end{tabular}
\end{center}
\caption{Ablation study on different pre-training objectives on ROCO dataset.}
\label{Ablation lrobjectives}
\end{table}

\section{Conclusion}
We introduce Uni-Mlip in this paper, a unified self-supervision framework tailored for enhancing medical vision-language pre-training. Our simple yet effective approach proves the advantage of integrating cross-modality, uni-modality, and fused-modality  self-supervision techniques. Additionally, our framework also unifies self-supervision from both the input level and the feature level. Moreover, our technique of freezing the mean and variance in the batch normalization layer is shown to be essential for integrating image-only contrastive learning in VLP on medical data. 

Extensive experiments validate the effectiveness of our method, as it achieves SOTA performance across various downstream medical tasks. Uni-Mlip not only holds promise for advancing vision-language pre-training in clinical applications but also presents potential benefits for domains characterized by limited multimodal data, such as satellite and agriculture imagery analysis.

\bibliography{main.bib}
\end{document}